\documentclass{article}

 \usepackage[sglblindworkshop]{neurips_2025}

\usepackage[utf8]{inputenc} 
\usepackage[T1]{fontenc}    
\usepackage{hyperref}       
\usepackage{url}            
\usepackage{booktabs}       
\usepackage{amsfonts}       
\usepackage{nicefrac}       
\usepackage{microtype}      
\usepackage{xcolor}         
\usepackage{graphicx}
\usepackage{hyperref}
\usepackage{url}
\usepackage{multirow}
\usepackage{enumitem}
\usepackage{amsmath}
\usepackage{natbib}

\usepackage{listings}
\usepackage{xcolor}

\title{DAOpt: Modeling and Evaluation of Data-Driven \\ Optimization under Uncertainty with LLMs}

\author{%
  WenZhuo Zhu\\
  Zhejiang University\\
  wzzhu@zju.edu.cn\\
  \And 
  Zheng Cui\\
  Zhejiang University\\
  zhengcui@zju.edu.cn\\
  \And
  Wenhan Lu\\
  Zhejiang University\\
  whlu@zju.edu.cn\\
  \AND
  Sheng Liu\\
  University of Toronto\\
  sheng.liu@rotman.utoronto.ca\\
  \And 
  Yue Zhao\\
  Peking University\\
  yzhao@phbs.pku.edu.cn\\
}

\begin{document}

\maketitle

\begin{abstract}
Recent advances in large language models (LLMs) have accelerated research on automated optimization modeling. While real-world decision-making is inherently uncertain, most existing work has focused on deterministic optimization with known parameters, leaving the application of LLMs in uncertain settings largely unexplored. To that end, we propose the DAOpt framework including a new dataset OptU, a multi-agent decision-making module, and a simulation environment for evaluating LLMs with a focus on out-of-sample feasibility and robustness. Additionally, we enhance LLMs' modeling capabilities by incorporating few-shot learning with domain knowledge from stochastic and robust optimization. 
\end{abstract}

\section{Introduction}
Formulating real-world decision-making problems into mathematical optimization models typically requires substantial expertise in the field of operations research (OR). Recent advances in large language models (LLMs) \citep{brown2020language,achiam2023gpt,zhu2024deepseek,guo2025deepseek} offer the potential to automate this process, translating textual problem descriptions into optimization formulations and executable solving code. Extensive work has focused on enhancing reasoning capability of LLM-based agents \citep{wei2022chain, zheng2023progressive,shinn2024reflexion}, built on which two main approaches have been proposed for automatic optimization modeling: (1) prompting methods such as Chain-of-Expert \cite{xiao2023chain} and OptiMUS \cite{ahmaditeshnizi2024optimus}
and (2) fine-tuning methods such as ORLM \cite{huang2024orlm}, LLMOpt \cite{jiang2025llmopt}, and OptiBench \cite{yang2025optibench}. 

Most existing work, however, has concentrated on {\it deterministic optimization} where the parameters are assumed known, emphasizing in-sample performance of the optimal decisions. Yet, practical decision-making is almost always carried out under uncertainty---arising from factors such as fluctuating demand or traffic conditions, where only partial information is available. Consequently, an optimal decision derived from the deterministic model may be over-optimistic, known as the ``optimizer's curse" \citep{smith2006optimizer}, or even become infeasible when implemented in practice \citep{birge1997introduction,ben2000robust}; see a detailed example in Appendix \ref{app: robust example}. 
The limitations of deterministic optimization have led to the development of {\it data-driven optimization under uncertainty} in OR, which explicitly accounts for uncertainty by leveraging observed data to manage risk in decision-making. This includes approaches such as stochastic optimization \citep{shapiro2021lectures}, robust optimization \citep{bertsimas2004price}, and distributionally robust optimization \citep{delage2010distributionally, mohajerin2018data, chen2020robust}. 

Despite the advantages demonstrated by LLMs in modeling deterministic problem, their role in automatically formulating and solving optimization problems under uncertainty remains largely unexplored. The main challenges arise in three dimensions:
(1) {\bf Mathematical reasoning}: unlike deterministic problems, uncertain settings require stronger reasoning ability of LLMs to construct algebraic models, identify stochastic elements, and incorporate historical data into the solution process; (2) {\bf Modeling expertise}: robust and stochastic optimization require deeper domain knowledge, such as calibrating uncertainty sets and deriving tractable reformulations via Lagrangian duality, which might be beyond the reliable capabilities of current LLMs; (3) {\bf Evaluation complexity}: deterministic settings only need comparison with a known ground truth, whereas data-driven settings require systematically assessing out-of-sample performance, emphasizing feasibility and robustness of the decision under uncertainty, posing greater experimental challenges. Nevertheless, we aim to advance research in this cross-disciplinary area by starting with two research questions:
\begin{itemize}[leftmargin=10mm,nosep]
\item How to evaluate the modelling ability of current LLMs in optimization under uncertainty?
\item How can their capability be enhanced by incorporating OR domain knowledge?
\end{itemize}

{\bf Contributions:} (1) To address the first question, we introduce a new dataset OptU and a simulation environment within the LLM-based multi-agent framework, enabling both in-sample and out-of-sample evaluations with a focus on feasibility and robustness rather than mere in-sample decision accuracy; (2) For the second, we leverage few-shot learning with domain knowledge from stochastic and robust optimization such as uncertainty set calibration and build-in Lagrangian duality operation, thereby strengthening the framework’s ability in optimization under uncertainty.

\section{DAOpt: An LLM-based Multi-agent Framework for Modeling and Evaluating of Data-driven Optimization under Uncertainty}
\label{gen_inst}

We aim to leverage LLMs to model data-driven optimization under uncertainty. In general, the problem can be formulated as follows:
\begin{equation}
\begin{array}{cl}
    \displaystyle \min_{x\in\mathcal{X}}  &\rho_0\left[f_0(\boldsymbol{x},\tilde{\boldsymbol{p}})\right]\\[2mm]
    s.t. & \rho_n\left[f_n(\boldsymbol{x},\tilde{\boldsymbol{p}})\right]\leq 0\quad  n=1,\cdots,N,
\end{array}
\end{equation}
where $\boldsymbol{x} \in \mathcal{X} \subseteq \mathbb{R}^{d_1}$ denotes the decision variables and $\tilde{\boldsymbol{p}} \in \mathbb{R}^{d_2}$ the random parameters. The function $f_n: \mathbb{R}^{d_1} \times \mathbb{R}^{d_2} \rightarrow \mathbb{R}$ is convex in $\boldsymbol{x}$ for fixed $\tilde{\boldsymbol{p}}$ and concave in $\tilde{\boldsymbol{p}}$ for fixed $\boldsymbol{x}$. The operator $\rho_n[\cdot]$ represents a risk functional, e.g., the expectation $\mathbb{E}_{\mathbb{P}}[\cdot]$ under distribution $\mathbb{P}$ of $\tilde{\boldsymbol{p}}$ and when $\mathbb{P}$ degenerates to a singleton, i.e., $\mathbb{P}[\tilde{\boldsymbol{p}} = \hat{\boldsymbol{p}}]=1$, the problem reduces to deterministic optimization. Different choices of $\rho$ correspond to difference modelling paradigms such as stochastic and robust optimization. Details of $\rho$ and associated formulations are provided in Appendix~\ref{app: Data-Driven Optimization Problems}.

\begin{figure}[ht]
\begin{center}
\includegraphics[width=1.0\textwidth]{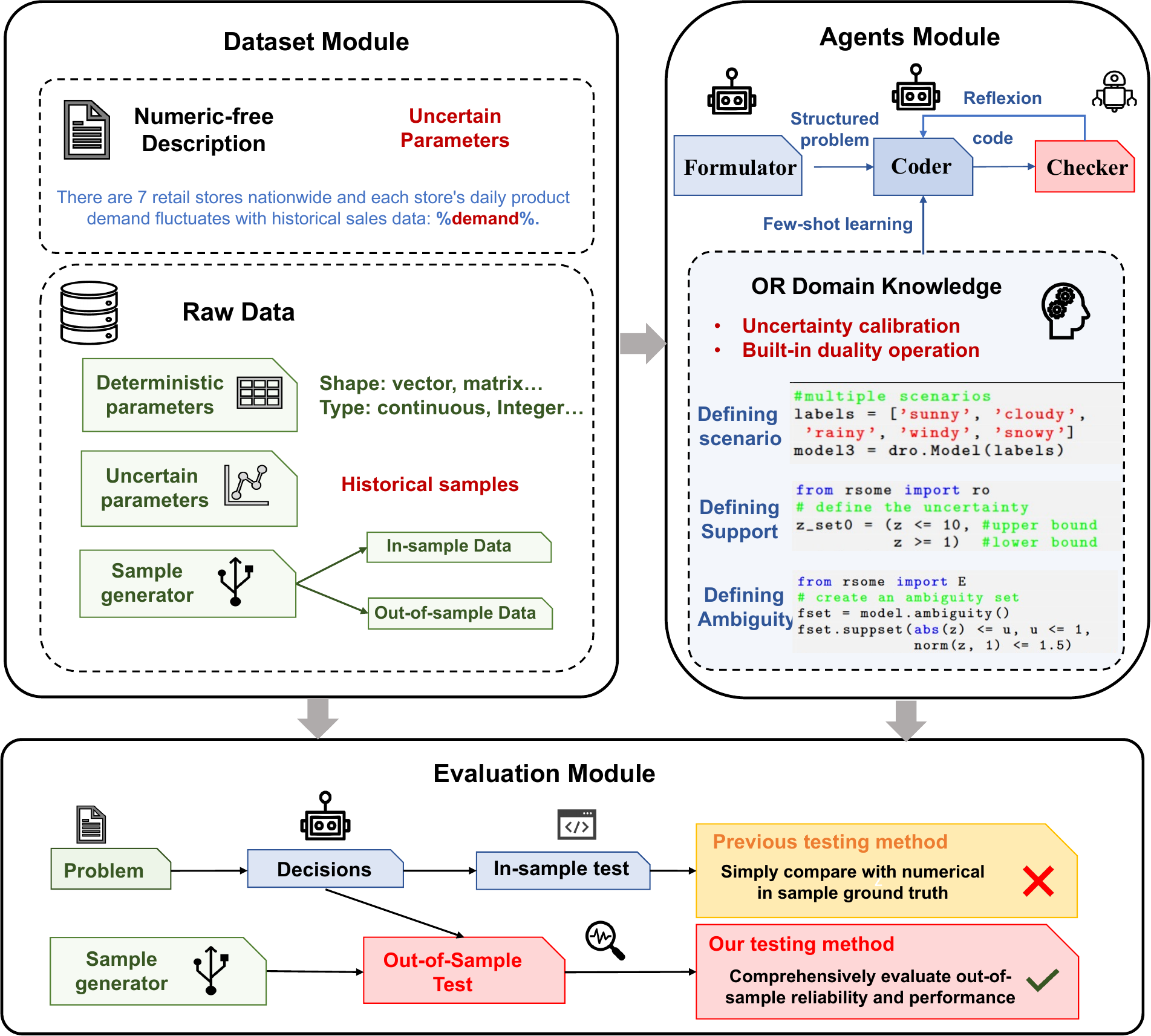}
\end{center}
\caption{DAOpt framework}
\label{fig: LLMRSOME system}
\end{figure}

Three main modules are designed to model and evaluate data-driven optimization problems within the DAOpt framework. The DAOpt framework is illustrated in Figure \ref{fig: LLMRSOME system}. Due to space limitations, the detailed explanations of these three modules are provided in Appendix \ref{app: DAOpt framework}. Below, we summarize the main characteristics of each module. 

{\bf 1. Dataset Module for Optimization under Uncertainty}. The OptU dataset is created to evaluate large language models (LLMs) in optimization under uncertainty. It consists of optimization problems converted into data-driven versions, utilizing LLMs with human oversight. Unlike existing datasets, OptU decouples problem descriptions from underlying data to enhance data privacy and scale with sample size. It includes historical records of problem parameters and incorporates a random sample generator to simulate diverse data environments. These features enable researchers to explore advanced modeling approaches, such as stochastic programming and robust optimization, making OptU particularly relevant for real-world applications in uncertain decision-making.

{\bf 2. Data-driven Decision-making Multi-agent Module}. This module uses a multi-agent system to automatically solve optimization problems from the dataset. It features three components: (1) the Data-driven Optimization Identifier, which converts natural language descriptions into structured models while identifying stochastic elements, (2) the OR Domain Knowledge Learner, which integrates the RSOME toolbox for formulating robust and stochastic optimization problems (an introdution to RSOME is provided in Appendix \ref{app: RSOME}), and (3) the Reflexion-based Checker, which ensures the correctness of the generated code. The framework is flexible, allowing for modification of components and fine-tuning of agents, ensuring an iterative, robust pipeline for turning natural language descriptions into actionable optimization solutions.

{\bf 3. Data-driven Simulation and Evaluation Module}. The evaluation process focuses on assessing the stability and performance of solutions under the out-of-sample data instances. Unlike deterministic optimization, where the solution is considered a ground truth, in data-driven optimization, no ground truth is available. The evaluation module measures out-of-sample feasibility, objective performance, and the deviation from the in-sample performance by testing the solution on independent out-of-sample instances. The out-of-sample testing ensures that the solution remains valid and reliable, even when faced with new data, reflecting the robustness and practical applicability of the model in real-world uncertain environments.

\section{Experiments and Discussion}
In this section, we assess the capability of our framework in modeling data-driven optimization problems using different optimization methods on general datasets. In Appendix \ref{app: stochastic inventory}, we provide a detailed comparison of the properties of solutions obtained through various optimization approaches leveraging large language models for a specific stochastic inventory network problem. The paper provides the data and code through the website \url{https://anonymous.4open.science/r/LLM-for-data-driven-optimization-problems-9528}.

In this framework, {\it in-sample data instances} refer to the data provided in the natural language descriptions, which are used to model and formulate the data-driven optimization problem. On the other hand, {\it out-of-sample data instances} are the independently generated data samples used in the evaluation to assess the real-world feasibility and performance of the optimal solutions. 

According to different modeling paradigms, our experiments involve the following optimization models: {\it robust optimization} {\bf (RO)}, where the uncertainty set is constructed based on the in-sample data instances; {\it distributionally robust optimization} {\bf (DRO)} with a Wasserstein ambiguity set; {\it deterministic optimization} {\bf (DM)} that uses the mean values of the in-sample data instances as parameters. We introduce the following key metrics for evaluation: 
\begin{itemize}
[leftmargin=10mm,nosep]
    \item 
    The \textit{successful rate of obtaining a valid solution} {\bf (SR)} represents the proportion of problems that are successfully programmed and solved to get a valid solution in the dataset.  
    \item  The \textit{out-of-sample feasibility rate} {\bf (FR)} defined as the proportion of out-of-sample test instances in which the solution satisfies all constraints of the problem.
    \item The {\it out-of-sample over-optimistic rate} {\bf(OpR)} measures the frequency with which a solution's performance on out-of-sample test instances is worse than its in-sample objective value.
\end{itemize} 
The detailed configurations of the optimization models and the formal definitions of the evaluation metrics are provided in Appendix \ref{app: metrics}.

We conduct experiments on our OptU dataset and evaluate the performance across different methods and optimization models. Specifically, we compare direct prompting, chain-of-thought (CoT) prompting, and our DAOpt framework using various base large language models. The dataset used for testing is constructed using seed OptU samples along with randomly generated samples. Each experiment is repeated three times for each base large language model, and the average performance is reported in Table \ref{tab: OptU datasets}.

\begin{table}[ht]
\centering
\caption{Experiments on OptU datasets: DeepSeek-V3 and GPT-4o}
\label{tab: OptU datasets}
\renewcommand{\arraystretch}{1.5}
\begin{tabular}{c|c|ccc|ccc|cc}
\hline\hline
 & Method     & \multicolumn{3}{c|}{Direct prompt} & \multicolumn{3}{c|}{CoT prompt} & \multicolumn{2}{c}{ DAOpt-LLM} \\ \cline{2-10}
& Model & DM      & RO        & DRO      & DM     & RO       & DRO     & RO              & DRO            \\ \hline
\multirow{3}{*}{DeepSeek-V3}   & SR $\uparrow$   & 0.36     & 0.36    & 0.39    & 0.39    & 0.44   & 0.36   & 0.33          & 0.31          \\
& FR $\uparrow$   & 0.27     & 0.47    & 0.41    & 0.32    & 0.51   & 0.23   & 0.63 & 0.71 \\ 
& OpR $\downarrow$ & 0.47     & 0.29    & 0.22    & 0.39    & 0.24   & 0.21   & 0.10 & 0.10 \\ \hline
\multirow{3}{*}{GPT-4o}       & SR $\uparrow$   & 0.11     & 0.29    & 0.42    & 0.17    & 0.33   & 0.42   & 0.36          & 0.31          \\
& FR $\uparrow$   & 0.27     & 0.57    & 0.33    & 0.40    & 0.61   & 0.40   & 0.68 & 0.80 \\ 
& OpR $\downarrow$ & 0.73     & 0.34    & 0.40    & 0.62    & 0.30   & 0.42   & 0.08 & 0.01 \\ \hline\hline
\end{tabular}
\end{table}
Our results show that the out-of-sample feasibility rate achieved by DAOpt is significantly higher than those from direct prompting and CoT prompting. Among all optimization models, the deterministic model yields the lowest out-of-sample feasibility rate—below 27\%—highlighting its inadequacy in uncertain environments. In contrast, the feasibility rate of the DRO model under DAOpt exceeds 70\%, demonstrating its robustness. Additionally, the RO and DRO models achieve much smaller out-of-sample over-optimistic rates, significantly lower than that of DM. This is because the robust models mitigate the optimizer's curse by safeguarding decisions against data perturbations and distributional shifts, rather than being overly adapted to the specific characteristics of the in-sample data.

Furthermore, DAOpt also attains a substantially higher successful rate than other methods when using GPT-4o. Although direct and CoT prompting occasionally achieve relatively high successful rates, the underlying LLMs do not rigorously solve robust optimization models through duality or robust counterpart techniques. In our experiment, these approaches often simply select a scenario perceived as "bad" and solve a deterministic problem, rather than formally deriving the dual reformulation. In contrast, our framework leverages the RSOME principles to accurately model and solve robust problems with strict mathematical duality.

\section{Conclusion}
In this paper, we presented the DAOpt framework for optimization under uncertainty, showcasing its potential to improve out-of-sample feasibility and robustness. Future work will focus on designing a multi-agent system that promotes further fine-tuning of each component, as well as integrating additional operations research (OR) domain knowledge to further enhance performance. Additionally, we plan to explore two-stage optimization problems, which are crucial in uncertain environments, to further advance the capabilities of LLMs in this domain.

\bibliographystyle{plain}
\bibliography{reference}


\appendix

\section{Example of Vulnerability in Deterministic Optimization Due to Data Uncertainty}\label{app: robust example}
\begin{figure}[!ht]
\centering
\includegraphics[width=0.9\textwidth]{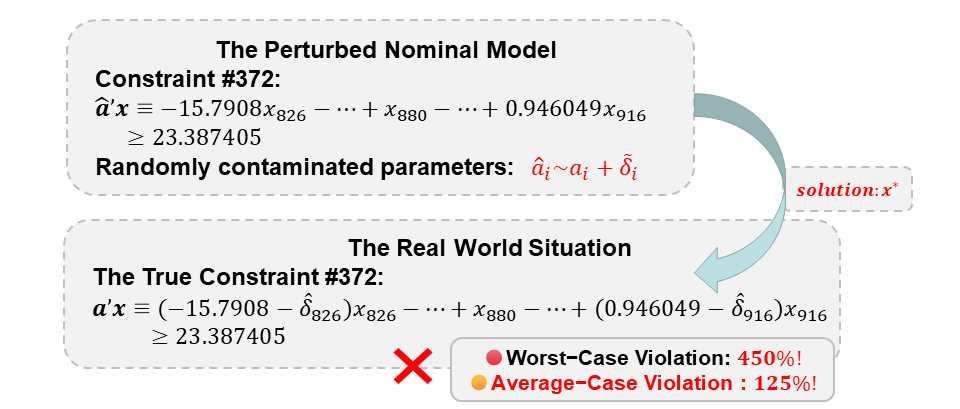}
\caption{Deterministic optimization can fail in an uncertain environment}\label{fig: Nemirovski_example}
\end{figure}

A key challenge in addressing optimization problems affected by uncertain data is that an optimal solution derived from nominal data may become severely infeasible in practical applications. Ben-Tal and Nemirovski (2000) demonstrated this phenomenon through an analysis of the PILOT4 problem from the well-known NETLIB collection. They examined constraint \#372 of the problem, assuming that its coefficients are at most 0.1\% error from the "true" values. In Figure \ref{fig: Nemirovski_example}, the "Perturbed Nominal Model" and the "Real World Situation" demonstrate how the solution derived from nominal data (represented by $\hat{a}$) can lead to a significant violation of the constraint in practice. The analysis revealed a striking vulnerability: in the worst-case scenario, the violation of this single constraint could be as large as 450\% of its right-hand side value. They also treated the uncertainty as random, with perturbations uniformly distributed in the range of [-0.001, 0.001] around the nominal coefficients. Even under this stochastic assumption, simulation results indicated that the mean relative violation of the constraint remained exceptionally high at 125\%.

\section{Data-Driven Optimization Problems (DDOPs)}\label{app: Data-Driven Optimization Problems}
{\bf Stochastic Optimization (SO)}: When $\rho[\cdot]:=\mathbb{E}_{\mathbb{P}}[\cdot]$, the decision maker is risk neutral, and the corresponding DDOPs minimize the expected objective function value. The main challenge of applying SO is that, in reality, we may not have access to the true distribution $\mathbb{P}$. 
Instead, there exist historical samples from the distribution, i.e.,  $\hat{\boldsymbol{p}}_i,$ $i\in [N].$ A common strategy is to evaluate the expectation using the empirical distribution $\hat{\mathbb{P}}_{N}:= \frac{1}{N}\sum_{i\in[N]}\delta_{\hat{\boldsymbol{p}}_i}$, which leads to classic sample average approximation schemes.

{\bf (Distributionally) Robust Optimization (DRO)}: When $\rho[\cdot]:=\sup_{\mathbb{P}\in\mathcal{P}}\mathbb{E}_{\mathbb{P}}[\cdot]$, we assume the decision maker is risk averse and the DDOPs minimize the worst-case expected loss when the distribution is drawn from an ambiguity set $\mathcal{P}$.
When $\mathcal{P}$ is a singleton $\{\mathbb{P}\}$, the problem reduces to SO. When all the elements of $\mathcal{P}$ are Dirac distributions, the problem reduces to Robust Optimization (RO).

{\bf Contextual Optimization (CO)}: 
When feature information is present, SO and DRO can be adapted to contextual optimization problems. In this setting, the uncertain parameters $\tilde{\boldsymbol{p}}$ may depend on observable contextual features $\boldsymbol{z} \in \mathbb{R}^{d_3}$, leading to a conditional distribution $\mathbb{P}_{\boldsymbol{p}|\boldsymbol{z}}$. Contextual Stochastic Optimization aims to minimize the conditional expectation given a context $\boldsymbol{z}$: $\rho[\cdot|\boldsymbol{z}]:=\mathbb{E}_{\mathbb{P}_{\boldsymbol{p}|\boldsymbol{z}}}[f(\boldsymbol{x}, \tilde{\boldsymbol{p}})]$.
This allows the decision variables $\boldsymbol{x}$ to dynamically adjust based on different contexts. When the conditional distribution is unknown, historical data $(\hat{\boldsymbol{z}}_i, \hat{\boldsymbol{p}}_i), i \in [N]$, is used to construct an empirical conditional distribution.
Similarly, Contextual Distributionally Robust Optimization goes one step further by considering the worst-case scenario under different contexts.


\section{DAOpt framework}\label{app: DAOpt framework}
\subsection{OptU: Dataset for Data-driven Optimization under Uncertainty}
We have constructed a new benchmark dataset designed to evaluate LLMs' capability in optimization under uncertainty, named OptU. The development of this dataset employed a human-AI collaborative framework, utilizing LLMs assisted by human oversight and curation. First, we collect representative deterministic optimization problems that are presented solely in natural language from existing datasets for deterministic optimization, e.g., NL4OPT, ComplexOR, IndustryOR and so on. Subsequently, we use LLMs to augment these problems by expanding the problem size and transforming the description into a data-driven version. All problems generated by the LLM then underwent a rigorous manual review process to ensure their validity. An example of our dataset and prompts used in our construction procedure is provided in Appendix \ref{app: example of dataset}.
Compared to existing datasets, our OptU dataset possesses the following characteristics that make it more reflective of real-world optimization problems:
\begin{itemize}[leftmargin=*,nosep]
    \item {\bf Decoupling of Problem Description and Data}. OptU explicitly decouples the problem description from the underlying data, ensuring that LLMs are only exposed to the descriptive context of the task. This approach enhances data privacy protection and is better aligned with data-driven contexts, where data volume scales with the number of samples. 

    \item {\bf Data-Driven Problem Parameters}. Unlike deterministic problems, OptU works with datasets that consist of historical records of problem parameters. This introduces a higher level of complexity and permits researchers to explore a more diverse range of modeling approaches, such as stochastic programming or robust optimization.
    \item{\bf Integrated sample generator}. OptU includes a flexible random sample generator capable of producing data samples with different statistical properties and underlying patterns based on preset parameters. The system is thus capable of simulating an extensive range of sophisticated, data-contingent environments.
\end{itemize}
\subsection{Decision Making Module}
In our decision-making module, we design a multi-agent system to automatically model and solve problems from the dataset based on given output criteria, natural language descriptions, and corresponding data. The final solution or decision is produced in a specified format. The module comprises the following three components:

\begin{itemize}[leftmargin=*,nosep]

\item {\bf Data-driven Optimization Identifier}. This agent is responsible for transforming unstructured natural language descriptions into a structured, formal representation. Unlike existing agents designed primarily for deterministic problem formulation, our agent must also identify stochastic components in the description, including uncertain parameters, probabilistic constraints, etc.
\item {\bf OR Domain Knowledge Learner}. Unlike deterministic optimization, stochastic and robust optimization problems admit multiple valid formulations, each tailored to different sources of uncertainty—a challenge even for human OR experts. Moreover, robust optimization problems require tractable reformulations derived via Lagrangian duality, which depend on both the problem structure and the choice of uncertainty set. Such derivations are difficult to expect reliably from LLMs. To address this issue, we integrate the RSOME toolbox \citep{chen2020robust,chen2023rsome}, which supports both stochastic and robust optimization and provides built-in duality-based reformulations. We further employ a few-shot prompting strategy to train the agent to interact with RSOME, enabling it to generate valid uncertainty-aware formulations automatically.
An introduction to the RSOME package is provided in Appendix \ref{app: RSOME}.
\item {\bf Reflexion-based Checker}. This component checks the code generated by the Coder and provides feedback on the result. If execution fails, the Checker returns relevant error information to the Coder for revision. If execution is successful, the module proceeds to output the solution.
\end{itemize}

This framework features a formulator and a coder as LLM agents with the coder possessing specialized knowledge in the RSOME package, while the checker is a program constructed through manual engineering. The framework allows flexible modification of components, supports the change of the base model of the agents and enables personalized further fine-tuning of each agent.
This tripartite architecture ensures a robust, iterative, and automated pipeline for transforming natural language problem descriptions into actionable decisions.

\subsection{Evaluation Module}
For data-driven problems, decision-makers are primarily concerned with obtaining solutions that are more stable and perform better under the true distribution, despite limited data information and the presence of perturbations. Consequently, the objective value derived from the nominal model in data-driven settings serves more as a prediction rather than a ground truth. This point shows the key difference between data-driven optimization and deterministic optimization. That is, in a data-driven setting, there is no such ground truth that can be solved from some model because the decision maker has no precise information about the true environment.

Once a reasonable solution/decision is obtained through the decision-making module, the out-of-sample testing module evaluates this decision using out-of-sample instances that simulate the true environment. We note that the testing samples are independently generated or distinctly sampled from the underlying distribution, such that they are completely separate from the samples used as input to the decision-making module. Under no circumstances should the decision-making module have access to any testing samples. We focus particularly on the following metrics:
\begin{itemize}[leftmargin=*,nosep]
   \item {\bf Out-of-sample feasibility}: Whether the obtained solution remains feasible under out-of-sample instances, reflecting the reliability of the solution. 
   \item {\bf Out-of-sample objective performance:} How the solution performs on the optimization objective under the out-of-sample instances (e.g., for cost minimization problems, this refers to the actual cost incurred in the true distribution).
   \item {\bf Prediction-related metrics:} The deviation of the objective value from the nominal model compared to that over the out-of-sample instances, i.e., whether the nominal prediction is conservative or optimistic.
\end{itemize}


\section{An example problem in OptU and prompts used in construction process}
\label{app: example of dataset}
\subsection{Structure of the dataset}
In this section, we introduce the structure of problems in our constructed dataset OptU.

Problems in our seed dataset consists of the following components:
\begin{itemize}
    \item {\bf description.txt} : This file stores the natural language description of a data-driven problem. Note that there is no data in the text. A symbol between \% means the data or historical sample of this parameter are stored in another file. 
    \item {\bf decision\_symbol.txt} : This document specifies the output format, i.e., the symbols that should be used for the decision of this problem, as well as the type—whether it is a scalar, vector, or matrix. If it is a vector or matrix, the dimensions are also defined. Only by adhering to the output format specified in this document can the out-of-sample module correctly read and test the decisions generated by the large language model.
    \item {\bf testing\_sample.json} : This file will later be adjusted by the random seed generator. This file stores the format of testing samples, the data of deterministic parameters and the means of the random parameters that will be later used to generate random input samples and testing samples. 
    \item {\bf truth.json} : This file stores the logic of testing the obtained decision, including the correct objective expression and constraints which are python logic expressions that be successfully executed by the checker if the variable of parameters are properly defined in the testing program.
    \item {\bf format\_description.json (Intermediate output of construction process)} : This is an auxiliary file containing the formalized expression of the problem that assists the random sample generator to correctly generate samples for each parameter which is uncertain and have historical data. Note that if the format in the testing samples are clear enough, it is not needed. This file should not be an input to the decision making module.
\end{itemize}

The above seed problem can not be directly used for testing as the samples are not properly generated. We will later append randomly generated input samples and testing samples to construct a sample problem. Through different setting of random sample generator, different kinds of samples can be generated to simulate different environments. The sample problem consists of following components:
\begin{itemize}
    \item {\bf description.txt}, {\bf decision\_symbol.txt}, {\bf truth.json} being exactly the same to the original seed problem.
    \item {\bf testing\_sample.json} regenerated by the random sample generator.
    \item {\bf training\_sample.json} which is the newly generated file that stores the data {\bf separated from the natural language description}. The file is named training\_sample.json because for the uncertain parameters some historical samples are generated as the input samples which will should be loaded to optimization model, and these input samples should be separated from the testing samples. For some deterministic parameters, the data should also be loaded from this file, representing a data-description separation scheme.
\end{itemize}

\subsection{An example problem in OptU}
Now we present an example problem of our OptU dataset:

(1) description.txt:
\begin{lstlisting}
As the head of logistics for the company, I need to develop an optimal product distribution plan to ensure sufficient supply for our stores while controlling transportation costs.

Current Situation
We manage multiple warehouses and stores:

Warehouses: The company has a total of 5 distribution centers, each with a certain amount of product inventory %inventory%. We must ensure that the quantity allocated to stores does not exceed actual inventory.

Stores: There are 7 retail stores nationwide, and each store's daily product demand fluctuates. Based on multiple periods of historical sales data: %demand%.

Transportation Costs: The shipping costs is randomly varying between different warehouses and stores, depending on factors such as distance and transportation method, Historical cost data is: 
%cost%

Optimization Objective
I need to determine how many products each warehouse should deliver to each store (X_{i,j}) to meet random demand while minimizing transportation costs. An operations research model will be established to solve this problem.
\end{lstlisting}

(2) decision\_symbol.txt:
\begin{lstlisting}
{
  "decision_variables": [
    {
      "symbol": "x",
      "meaning": "Number of products delivered from each warehouse to each store",
      "shape": [5, 7]
    }
  ]
}
\end{lstlisting}
(3) truth.json:
\begin{lstlisting}
{
  "constraints": [
    "np.sum(np.asarray(x), axis=1) <= np.asarray(inventory)",
    "np.sum(np.asarray(x), axis=0) >= np.asarray(demand)",
    "np.asarray(x) >= 0"
  ],
  "objective": "np.sum(np.asarray(cost) * np.asarray(x))",
  "problem_type": "min"
}
\end{lstlisting}

(4) testing\_sample.json \& training\_sample.json 
\begin{lstlisting}
{
  "sample_size": 5,
  "parameters": [
    {
      "symbol": "inventory",
      "meaning": "Inventory at each warehouse",
      "is_random": 0,
      "value": [84, 35, 50, 40, 200],
      "sample": null,
      "type": "Integer",
      "shape": [5],
      "is_non_negative": 1
    },
    {
      "symbol": "demand",
      "meaning": "Average demand at each store",
      "is_random": 1,
      "value": [32.2, 32.4, 21.8, 20.6, 36.8, 42.0, 24.6],
      "sample": [
        [35.0, 30.0, 18.0, 22.0, 40.0, 45.0, 26.0],
        [30.0, 35.0, 25.0, 19.0, 35.0, 40.0, 23.0],
        [33.0, 32.0, 20.0, 21.0, 38.0, 43.0, 25.0],
        [29.0, 34.0, 22.0, 20.0, 37.0, 44.0, 24.0],
        [34.0, 31.0, 24.0, 21.0, 34.0, 38.0, 25.0]
      ],
      "type": "Continuous",
      "shape": [7],
      "is_non_negative": 1
    },
    {
      "symbol": "cost",
      "meaning": "Average transportation cost from each warehouse to each store",
      "is_random": 1,
      "value": [[4.2, 4.8, 4.1, 4.9, 4.3, 4.9, 4.3], [4.3, 4.9, 4.9, 3, 4.9, 4.0, 6.0], [4.9, 4.7, 2, 4, 4.8, 4.0, 4.7], [4.7, 4.9, 5.0, 7, 4.3, 4.1, 4.6], [4.7, 4.9, 6, 4.8, 4.7, 8, 4.3]],
      "sample": [
        [[7.5, 7.8, 7.2, 0.8, 1.2, 1.9, 7.8], [7.4, 2.7, 5.1, 5.8, 2.6, 4.8, 3.4], [6.1, 7.9, 7.3, 2.1, 7.3, 1.1, 7.2], [8.1, 7.0, 1.8, 1.0, 2.0, 8.3, 10.0], [3.4, 8.5, 8.3, 3.6, 7.0, 6.1, 2.0]],
        [[6.7, 8.8, 4.8, 1.0, 1.1, 1.8, 7.7], [8.2, 3.2, 5.1, 5.4, 3.4, 5.2, 2.5], [5.3, 7.4, 7.8, 2.0, 9.4, 0.8, 7.2], [7.9, 7.3, 2.1, 1.1, 2.4, 9.7, 9.3], [3.8, 9.2, 8.4, 4.1, 6.7, 6.7, 1.7]],
        [[6.3, 8.6, 6.0, 1.2, 0.8, 1.9, 8.6], [7.9, 3.2, 4.6, 5.2, 2.8, 4.4, 3.2], [5.0, 8.2, 6.7, 1.9, 7.7, 1.0, 6.3], [7.1, 7.9, 2.1, 1.0, 1.7, 8.7, 8.8], [4.3, 9.5, 8.6, 3.5, 7.5, 4.9, 2.2]],
        [[5.2, 9.8, 6.7, 1.1, 1.0, 2.5, 7.9], [6.7, 2.8, 4.5, 6.6, 2.5, 5.0, 2.4], [4.2, 6.9, 7.4, 2.0, 8.6, 1.0, 6.5], [8.2, 6.7, 1.6, 1.0, 1.8, 9.9, 9.2], [3.9, 9.4, 7.7, 4.5, 9.0, 5.9, 2.0]],
        [[6.8, 7.9, 5.4, 1.0, 1.2, 1.9, 7.1], [7.7, 2.9, 4.7, 6.8, 3.4, 5.6, 3.0], [4.6, 9.7, 8.7, 1.6, 6.9, 0.9, 6.7], [9.1, 6.5, 2.3, 1.1, 2.1, 9.4, 10.0], [3.5, 9.2, 7.9, 4.6, 8.6, 6.7, 2.2]]
      ],
      "type": "Continuous",
      "shape": [5, 7],
      "is_non_negative": 1
    }
  ]
}
\end{lstlisting}

These two files have different sample sizes and different generated samples.
The training\_sample.json should only be used by the decision making model to out a decision. The testing\_sample.json should only be used by the out of sample test module to evaluate performance.

\subsection{Prompts used in dataset construction}

(1) Prompts for transforming deterministic descriptions into uncertain ones:
\begin{lstlisting}
You will be given a description of a deterministic problem and your task is to generate a new version of the description which is a data driven one(e.g. the parameters are unknown and only historical data are observed). You should follow these rules: 
-Identify uncertain parameters as many as possible in the problem description. 
-For uncertain parameter(s) you identified , use %symbol% to replace their data, where the symbol between % is the symbol denoting the uncertain parameters. You should not use %symbol% to replace other parameters like the size.
-Expand the problem size and as much as possible. Try your best to make sure both the objective and constraints involve uncertain parameters. The expanded size should be a number, not a symbol
-Only show that some parameters are uncertain and there are historical samples about them, do not add anything unnecessary.
Now the original description is:

{problem}

Now give only the new description of the problem and the explanation should be omitted.
\end{lstlisting}
After the large language model gives the transformed description, the human checkers check following things: if the randomization of parameters is reasonable, if the expanded parameters and decisions are reasonable, reorganize the decisions and decisions as vector forms and matrix forms if necessary, if the long data in the description is still numerical instead of being a symbol, if the generated new description should be discarded.

(2) Prompts of generating structured description of the problem. 
\begin{lstlisting}
You will be given a description of a real-world problem. Your task is to understand the real-world problem and format an optimization model according to the description. You need to follow these rules:
- Recognize decision variables and parameters in the problem.
- Identify the types (Continuous, Integer or Binary) and the shapes of the decision variables.
- Identify whether these decision variables should be non-negative.
- Identify whether these parameters are random or deterministic.
- Identify the objective of the problem.

The result should be summarized in the json format described as below:
- It should contain four aspects: "parameters", "decision variables", "description", "objective"

- Each parameter consists of eight attributes: "symbol", "meaning", "is_random", "value", "sample", "type", "shape", "is_non_negative".
- The attribute "symbol" should be readable. For vector and matrix form parametrs, "symbol" should not involve subscript like "_j","_i","_k" etc.
- If the parameter is random, the value of the attribute "is_random" should be 1. Furthermore, if historical samples are provided, the attribute "value" should be strictly the average of these samples and the attribute "sample" should be a list of these samples formatted as [ , , , ]. If only symbols are provided and organized as %parameter_symbol%, only the string between '%' is needed., fill in the "value" and "sample" with the symbol. Otherwise, attributes "value" and "sample" should be null.
- If the parameter is deterministic, the value of the attribute "is_random" should be 0. Furthermore, the attribute "value" should be the value of the parameter and the attribute "sample" should be null.
- for "type", "shape", "is_non_negative", refer to the following instruction for decision variables.

- Each decision variable consists of five attributes: "symbol", "meaning", "type", "shape", "is_non_negative".
- The attribute "symbol" should be readable. For vector and matrix form decisions, "symbol" should not involve subscript like "_j","_i","_k" etc.
- For the decision variables, the attribute "type" should have value "Continuous", "Integer" or "Binary" according to the type of the decision variable which you need to carefully identify from the description! 
- For scalar decision variables, the attribute "shape" should be empty list []. For n-dimensional vector decision variables, the attribute "shape" should be a list [n] with one element. For n\times m-dimensional matrix decision variables, the attribute "shape" should be a list [m,n] with two elements.
- If the decision variable is non-negative, the attribute "is_non_negative" should be 1, otherwise 0.

- For the description, it should be the original description with the value of the parameters replaced by their attribute "symbol". Also omit the paramters' unit of measurement.


Now the description is:
\n{problem}\n 
Give only the json format and omit other information.

\end{lstlisting}

\begin{lstlisting}
Ok, now generate some reasonable values for the empty parameters in the 'value' attribute based on the shape, but do not change the 'sample' attribute, because this step is to generate the mean of the sample. I will generate the actual sample by calling a random number generator based on the 'value' later.


\end{lstlisting}

After the structured description is obtained from the large language model. The human checkers check if the generated structured description correctly identifies all the deterministic parameters, all the uncertain parameters and they types.

Then fill in the parameters' values using prompts and check if the parameters' values are properly set to make the problem valid and feasible.

(3) Prompt of generating testing\_sample.json
\begin{lstlisting}
You are an mathematical modeling expert. You will be given a problem formated as a json file. Your task is to edit the json file and give a new json file. You need to follow these rules: -Add "sample_size" attribute in the first. 
-The "sample_size" need to first be generated randomly in the [20,100]
-Keep "parameters" part unchanged, it should follow exactly the format in the original json file. -Omit other attributes in the original json file.
Now the json file is:
\n{problem}\n 
Give only the .json format and omit other information .
\end{lstlisting}
After testing\_sample.json is generated, the human checkers need to check if the json format is correct and loadable. The human checkers also need to check if the symbols of parameters in this json file match the symbol provided in the description.

(4) Prompt of constructing decision\_symbol.txt
\begin{lstlisting}
You are an mathematical modeling expert.
You will be given a problem formated as a json file. Your task is to reduce this json file, leaving only its decision_variables as well as their symbol,desscription and shape attributes. The format of the output should follow the original one. Now the content of the json file is:
\n{problem}\n
Give only the .json format of the result and omit other information.
\end{lstlisting}

(5) Prompt of constructing truth.json:
\begin{lstlisting}
You are an mathematical modeling expert and some knowledge about python syntax is required.

You will be given a problem formated as a json file. Your task is to identify all the constraints and the objective and present the result in a json format.

You need to follow these rules:
-Your result should be a json file formulating the constraint, objective and type of the problem.
-The constraints should be a list organized as a list, each one should be an string of the math expression using exactly the symbol provided by the reference.
-The constraints should involve in-explicit ones like non-negativity and decision variable's type. The type constraints of decision variables with symbol(symbol) should be described as: is_integer(symbol), is_binary(symbol).
-Each constraint should function as a code representing logical expression (can be vector or matrix)
-The objective should be a string of the math expression with exactly the symbol provided by the reference json file. Strictly follow only the symbols of the parameters and decision variables and remember that min or max should never appear in it!
-The type should be min or max denoting the problem type.
-In your constraints and objective involving matrixes and vectors data you need to transform these data into np.array use np.asarray(). you need to use numpy like np.sum() to better describe your constraints. vector constraints are supported.

Now the json file is:
\n{problem}\n
Give only the .json format and omit other information .

\end{lstlisting}

\begin{lstlisting}
Write code, fill in the data yourself. Strictly use raw logical expressions to test constraints. Directly execute each constraint expression from the JSON to verify syntax and logic, and test whether it can run properly.
\end{lstlisting}
After truth.json is generated by the large language model, the human checkers need to check if the json format is correct, if the constraints and objective are correctly defined, if every element in this json file is executable. The human checker can also prompt the large language model to produce a code for testing if each constraint is an executable logical expression in python.

\section{Introduction to the RSOME package}\label{app: RSOME}

RSOME -- Robust Stochastic Optimization Made Easy ( \url{https://xiongpengnus.github.io/rsome/}) -- is an open-source Python package for modeling generic optimization problems. 
Models in RSOME are constructed by variables, constraints, and expressions that are formatted as N-dimensional arrays. 
These arrays are consistent with the NumPy library in terms of syntax and operations, including broadcasting, indexing, slicing, element-wise operations, and matrix calculation rules, among others. In short, RSOME provides a convenient platform to facilitate the development of optimization models and their applications.
This package can model Stochastic and robust optimization problems in a unified way; using this package only requires formulating the model following specific syntax (language). 
The package automatically does the reasoning part, e.g., deriving robust counterparts or dual models.

\section{Setup of the numerical study} \label{app: metrics}

According to different optimization paradigms, our experiments involve the following optimization models: 
\begin{itemize}
\item {\bf RO:} Robust optimization model, where the uncertainty set is defined by the lower and upper bounds of the uncertain parameters based on the in-sample data instances. 
\item {\bf DRO-0/0.1/0.5:} Distributionally robust optimization model with a Wasserstein ambiguity set, with base radius 0/0.1/0.5. The true radius is the base radius scaled by the parameter values, and the reference distribution is the empirical distribution from the in-sample data instances. 
\item {\bf DM:} Deterministic model that uses the mean values of the in-sample data instances as parameters.
\end{itemize}

We introduce the following key metrics for evaluation:
\begin{itemize}
    \item{\bf SR}: The \textit{successful rate of obtaining a valid solution}. This metric represents the proportion of problems that are successfully programmed and solved to get a valid solution in the dataset.  A problem is considered successful if the generated code is successfully executed and gives a decision of the required format.
    \item{\bf FR}:
    The \textit{out-of-sample feasibility rate}, defined as the proportion of feasible scenarios in the out-of-sample instances:
    $
    \text{FR} = \frac{
        |\mathcal{N}_{\text{feas}}|
    }{
        N_{\text{out}}
    }.
    $ When considering the entire dataset, the feasibility rate is the average of each successful problem's feasibility rate.

    \item{\bf Obj}: 
    The \textit{average out-of-sample objective value for feasible out-of-sample instances}, computed as:
    $
    \text{Obj} = \frac{1}{|\mathcal{N}_{\text{feas}}|} \sum_{i \in \mathcal{N}_{\text{feas}}} v_i^{\text{out}}.
    $
    where \( \mathcal{N}_{\text{feas}} \) is the index set of feasible out-of-sample instances, \( v_i^{\text{out}} \) is the objective value of sample \( i \).


    \item{\bf OpR}:
    The \textit{out-of-sample optimistic rate}, representing the proportion of feasible out-of-sample instances where the out-of-sample performance is worse than the in-sample objective:
   $
   \text{OpR} = \frac{1}{|\mathcal{N}_{\text{feas}}|} \sum_{i \in \mathcal{N}_{\text{feas}}} 
      \mathbb{I}(v_i^{\text{out}} > v^{\text{in}})
   $, where \( \mathbb{I}(\cdot) \) is the indicator function.
\end{itemize}

\section{Experiments on a stochastic inventory network problem}\label{app: stochastic inventory}
We conduct experiments on different data-driven models on an inventory network problem and evaluate their out-of-sample performances. We set the data-generating distribution to be an independent lognormal distribution with a coefficient of variation of $0.3$. The in-sample instance size is set to $50$ and the size of out-of-sample instance is set to $10^3$ to simulate the true distribution. The samples are generated using $10$ different random seeds and the average performance is reported in Table \ref{tab: inventory}.

\begin{table}[ht]
\centering
\caption{Experiments on a stochastic inventory network problem}
\label{tab: inventory}
\begin{tabular}{c|cccc|cc}

Method     & \multicolumn{4}{c|}{DOpt-LLM}   & \multicolumn{2}{c}{w.o. Domain knowledge} \\ \hline
\multicolumn{1}{c|}{Model} & \multicolumn{1}{c}{RO} & \multicolumn{1}{c}{DRO-0.5} & \multicolumn{1}{c}{DRO-0.1} & \multicolumn{1}{c|}{DRO-0} & RO & DM             \\ \hline
FR                            & 0.89                  & 0.62                        & 0.89                        & 0.77                       & 0.56                 & 0.02 \\
Obj                             & 1131                & 1066             & 1125                      & 1120                     & 1099               & 414                               \\
OpR                             & 0.00                     & 0.02                         & 0.00                           & 0.48                       & 0.30                 & 0.47                                \\ \hline
\end{tabular}

\end{table}

We notice that the out-of-sample feasibility rates of deterministic models fail to exceed 3\%. That is, in a complex data-driven problem setting, the solution obtained from the deterministic model can barely meet the requirements.
Meanwhile, we see from Table \ref{tab: inventory} that the robust and distributionally robust models implemented by the RSOME package achieve a high feasibility rate and better than the RO model without domain knowledge.

The out-of-sample cost of the solution produced by the deterministic model is unrealistically low because the deterministic model only considers one scenario and completely neglects most other possible realizations of parameters. The DRO model with Wasserstein ambiguity set typically achieves lower out-of-sample objective when the radius is properly set. That is because DRO model ensures feasibility and stability against uncertainty in unseen data, rather than being overly tailored to the specific training data.


\end{document}